\documentclass{article}

\usepackage{authblk}
\usepackage[letterpaper,top=2cm,bottom=2cm,left=2.5cm,right=2.5cm,marginparwidth=1.75cm]{geometry}

\usepackage{amsmath,amssymb, bm}
\usepackage{graphicx}
\usepackage[colorlinks=true, allcolors=blue]{hyperref}

\title{Learning Reservoir Dynamics with Temporal Self-Modulation}
\author[1,2]{Yusuke Sakemi}
\author[1,5,6]{Sou Nobukawa}
\author[3]{Toshitaka Matsuki} 
\author[4]{Takashi Morie}
\author[1,2]{Kazuyuki Aihara}

\affil[1]{Research Center for Mathematical Engineering, Chiba Institute of Technology, Narashino, Japan}
\affil[2]{International Research Center for Neurointelligence (WPI-IRCN), The University of Tokyo, Tokyo, Japan}
\affil[3]{Oita University, Oita, Japan}
\affil[4]{Graduate School of Life Science and Systems Engineering, Kyushu Institute of Technology, Kitakyushu, Japan}
\affil[5]{Department of Computer Science, Chiba Institute of Technology, Narashino, Japan}
\affil[6]{Department of Preventive Intervention for Psychiatric Disorders, National Institute of Mental Health, National Center of Neurology and Psychiatry, Tokyo, Japan}

\begin{document}
\maketitle

\begin{abstract}
Reservoir computing (RC) can efficiently process time-series data by transferring the input signal to randomly connected recurrent neural networks (RNNs), which are referred to as a reservoir. 
The high-dimensional representation of time-series data in the reservoir significantly simplifies subsequent learning tasks. 
Although this simple architecture allows fast learning and facile physical implementation, 
the learning performance is inferior to that of other state-of-the-art RNN models. 
In this paper, to improve the learning ability of RC, we propose self-modulated RC (SM-RC), which extends RC by adding a self-modulation mechanism. 
The self-modulation mechanism is realized with two gating variables: an input gate and a reservoir gate. 
The input gate modulates the input signal, and the reservoir gate modulates the dynamical properties of the reservoir. 
We demonstrated that SM-RC can perform attention tasks where input information is retained or discarded depending on the input signal. 
We also found that a chaotic state emerged as a result of learning in SM-RC. This indicates that self-modulation mechanisms provide RC with qualitatively different information-processing capabilities. 
Furthermore, SM-RC outperformed RC in NARMA and Lorentz model tasks. In particular, SM-RC achieved a higher prediction accuracy than RC with a reservoir 10 times larger in the Lorentz model tasks. 
Because the SM-RC architecture only requires two additional gates, it is physically implementable as RC, providing a new direction for realizing edge AI.
\end{abstract}

\section*{Introduction}

Vast amounts of data are generated and observed in the form of time series in the real world. 
Efficiently processing these time-series data is important in real-world applications such as forecasting the renewable energy supply and monitoring sensor data in factories. 
In the past decade, data-driven methods based on deep learning have progressed significantly and have successfully linked data prediction and analysis to social values, and they are becoming increasingly important \cite{Ismail2019deep,Dong2021survey}.
However, the computational load of data-driven methods results in considerable energy consumption, limiting their applicability \cite{Thompson2020computational, Patterson2021carbon}.
In particular, to perform prediction and analysis near the location where the data are generated, which is called edge AI, a high energy efficiency is required \cite{Murshed2021machine}. 

Reservoir computing (RC) is attracting attention as a candidate for edge AI because it achieves high prediction performance and a high energy efficiency.
The RC model consists of an input layer, a reservoir layer, and an output layer \cite{Jaeger2001echo,Maass2002realtime}. 
The reservoir layer is typically a recurrent neural network (RNN) with fixed random weights.
Because only the output layer is usually trained in RC, the training process is faster than those of other RNN models such as long-short term memory (LSTM) \cite{Hochreiter1997long} and gated recurrent units (GRUs) \cite{Cho2014learning}.
In addition, because the reservoir layer can be configured with various dynamical systems \cite{Dambre2012information}, 
its high energy efficiency has been demonstrated through physical implementations \cite{Tanaka2019recent, Nakajima2020physical}.

However, as only the output layer is trained in RC, it is unclear whether it can achieve comparable prediction accuracy for real-world applications to other state-of-the approaches \cite{Vlachas2020backpropagation}. 
To improve the computational power of RC, various RC architectures and methods have been proposed \cite{Sum2020review}.
Recent proposals include structures that combine convolutional neural networks \cite{Tong2018reservoir}, parallel reservoirs \cite{Pathak2018model,Kawai2022self}, multilayer (deep) RC \cite{Gallicchio2017deep}, 
methods that use information from past reservoir layers \cite{Sakemi2020model}, and
regularization methods that combine autoencoders \cite{Chen2020autoreservoir}.
These studies indicated that the performance can be improved by using an appropriate reservoir structure.
However, because only the output layer is trained in these methods, the performance improvement is limited.

One promising approach for improving the flexibility of information processing in RC is to temporally vary the dynamical properties of the reservoir layer to adapt to the input signal.
An architecture that feeds the output back to the reservoir can realize this \cite{Sussilo2009generating, Rivkind2017local}.
Sussilo and Abbott proposed a FORCE learning method for stably training an RC model with a feedback architecture and reported that the network can learn various types of autonomous dynamics \cite{Sussilo2009generating}.
This architecture has been extended to spiking neural networks \cite{Nicola2017supervised} and applied to reinforcement-learning tasks \cite{Maxtsuki2019adaptive}.
However, although these feedback connections are thought to control the dynamics according to tasks \cite{Rivkind2017local}, their control is limited because the connections are random.

Research has also been conducted to acquire task-dependent dynamics in the reservoir layer by training not only the output layer but also the reservoir layer.
Intrinsic plasticity \cite{Schrauwen2008impriving} is a method for making the outputs of neurons closer to a desired distribution, and Hebbian rule or anti-Hebbian rule \cite{Yusoff2016modeling} allows control of correlations between neurons.
These methods increased the prediction accuracy and memory capacity \cite{Morales2021unveiling}.
Lage and Buonomano proposed innate training, which realizes a long-term memory function by training some connections in the reservoir layer to construct an attractor that is stable for a certain period of time \cite{Laje2013robust}.
Inoue et al. used this method to construct a chaotic itinerary \cite{Inoue2020designing}.
The aforementioned studies demonstrated that it is possible to perform tasks that are difficult to achieve with conventional RC models by training the reservoir layer.
However, in all the methods used, the properties of the trained reservoir layer were static (e.g., fixed network connections in time), limiting the diversity of the dynamics.

Recently, the attention mechanism has been considered as an effective method for realizing information processing adapted to the input signal.
In deep learning, the introduction of the attention mechanism was one of the breakthrough techniques proposed in recent years \cite{Bahdanau2014neural, Vaswani2017attention}.
The introduction of this mechanism allows efficient learning through the selection and processing of important information and has impacted various research fields, such as natural language processing \cite{Han2022survey, Jumper2021highly}.
In neuroscience, attention is considered an important factor for realizing  cognitive function, and neuromodulation is known as a neural mechanism closely related to attention \cite{Thiele2018neuromodulation}.
Neuromodulation is caused by neurons in brain areas such as the basal forebrain releasing neuromodulators such as acetylcholine into various brain areas to modulate the activity of neurons therein \cite{Thiele2018neuromodulation, Edelmann2018dopaminergic, Palancios2019neuro}.
The attention mechanism can increase the efficiency of information processing.
However, in RC, the input signal is uniformly transferred to the reservoir layer and converted into high-dimensional features; thus, there is no attention mechanism.

\begin{figure*}
\centering
\includegraphics[clip, width=\textwidth]{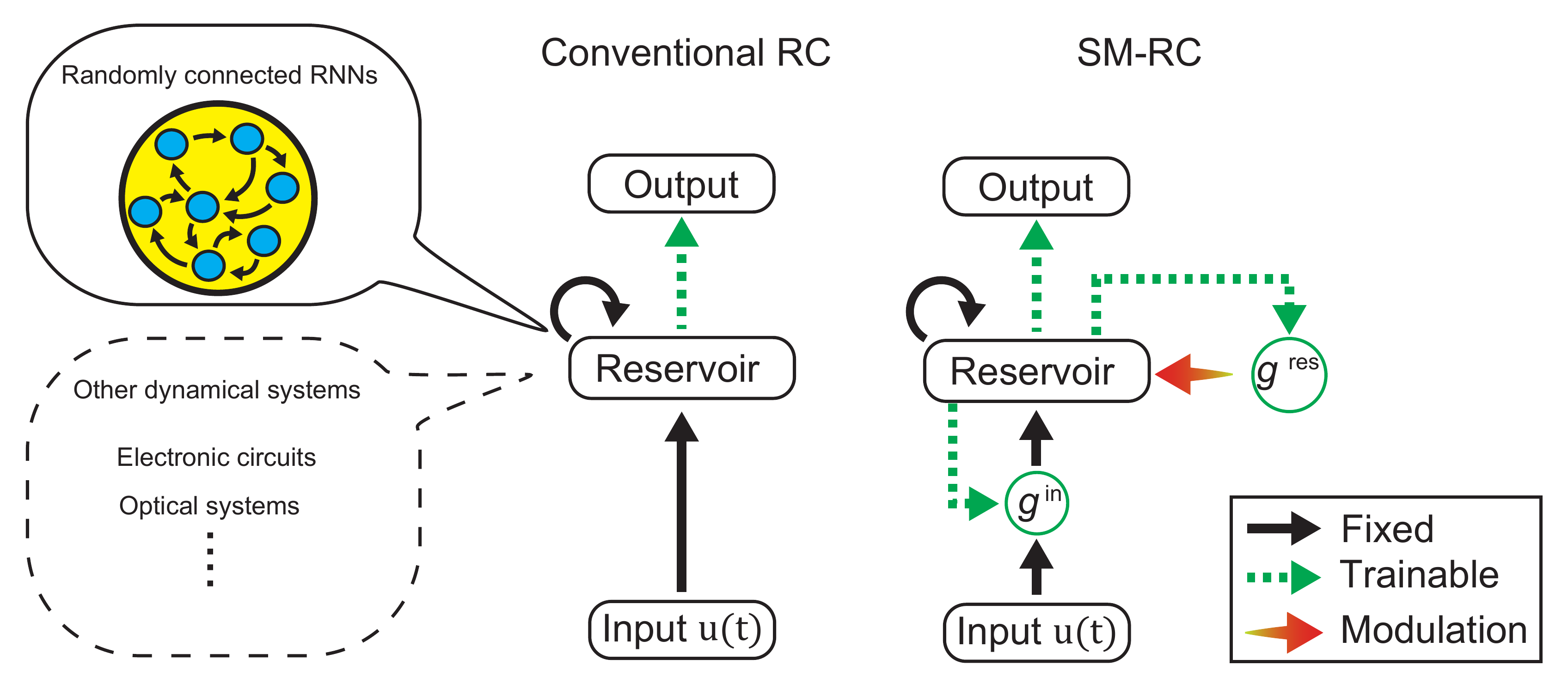}
\caption{Comparison of the conventional RC and SM-RC architectures. 
In conventional RC, the input signal is directly transferred to the reservoir layer, and the output is obtained from the reservoir layer. 
The reservoir layer is typically constructed with an RNN having fixed weights, but various dynamical systems can be used as reservoirs, which can be implemented with electronic circuits, optical systems, and other physical systems. 
SM-RC inherits the basic architecture of conventional RC, including the reservoir layer and output layer. 
However, the input signal is modulated by the input gate $g^\text{in}$ before being transferred to the reservoir. 
Additionally, the reservoir dynamics are modulated with the reservoir gate $g^\text{res}$. 
Both gates are controlled by the feedback from the reservoir layer. 
The black solid arrows and green dotted arrows represent the randomly initialized fixed connections and trainable connections, respectively. 
The red gradient arrow represents the function that changes the dynamical properties of the reservoir. 
}
\label{fig:proposed}
\end{figure*}

In this paper, we propose self-modulated RC (SM-RC) an architecture that incorporates the advantages of the above feedback structure, reservoir-layer learning, and the attention mechanism.
SM-RC has trainable gates that can dynamically modulate the strength of input signals and the dynamical properties of the reservoir layer.
Thus, it is possible to learn the reservoir dynamics adapted to the input signal, allowing information processing, e.g., attention, and significantly improving the learning performance in a wide range of tasks. 
Importantly, the gate structure has at most two variables and is controlled by feedback from the reservoir layer; thus, we expect SM-RC to be highly hardware-implementable, similar to conventional RC.
In the following, we focus on the learning performance and model characteristics of SM-RC.
In particular, we compare the prediction performance with that of conventional RC through simple attention tasks, NARMA tasks, and Lorentz model tasks.
We also discuss prospects for hardware implementation.

\section*{Results}

Figure \ref{fig:proposed} shows the SM-RC architecture.
In addition to the input layer, reservoir layer, and output layer, 
the proposed architecture has an input gate $g^\text{in}$ that modulates the input signal and a reservoir gate $g^\text{res}$ that changes the dynamical properties of the reservoir layer, both of which are controlled by feedback from the reservoir layer.

These properties significantly extend the functionality of conventional RC,
wherein input information is ``uniformly'' transferred to the reservoir layer.
In addition, the information of an input signal stored in the reservoir layer tends to decay over time, which is related to the echo-state property and fading memory \cite{Yildiz2012revisiting,Dambre2012information}.
These properties of conventional RC do not pose a significant problem for relatively simple tasks, but they severely limit the regression performance for tasks that involve capturing long or complex temporal dependencies \cite{Sakemi2020model}. 
SM-RC overcomes the shortcomings of conventional RC by introducing the self-modulation mechanism.

In this study, for simplicity, we construct SM-RC on the basis of the echo-state network (ESN) \cite{Jaeger2001echo, Lukosevicius2012practical}, which is the most commonly used form of RC.
The ESN is a discrete-time dynamical system, and the reservoir layer consists of an RNN with fixed weights.
In addition, we consider an input gate that dynamically changes the input strength and a reservoir gate that dynamically changes the internal connection strength in the reservoir layer.
In this case, the input $u^\text{res}$ to the reservoir layer and the spectral radius $\rho ^\text{res}$ of the reservoir layer are time-modulated as follows:
\begin{flalign}
\bm{u}^\text{res}(t) &= g^\text{in}(t-1) \bm{u}(t), \\
\rho ^\text{res}(t) &= g^\text{res}(t-1) \hat{\rho} ^\text{res}, \label{eq:reservoir_gate}
\end{flalign}
where $\bm{u}(t)$ is the original input vector and $\hat{\rho} ^\text{res}$ is the spectral radius of the inner connection matrix of the unmodulated reservoir layer.
Note that $g^\text{in}$ and $g^\text{res}$ are trained scalar gates.
From the function of these gates, we can intuitively understand how SM-RC extends RC.
For example, the input gate can realize attention by sending important input signals to the reservoir layer and discarding other information, and
the reservoir gate can change the memory retention characteristics depending on the input signal.

To evaluate the learning performance of SM-RC from various viewpoints, we compared it with that of conventional RC for three tasks with different characteristics: simple attention tasks, NARMA tasks, and Lorentz model tasks. 
In all the experiments, the number of neurons in the SM-RC reservoir layer $N^\text{res}$ was set as 100, and the hyperparameters were fixed to the same values for SM-RC. 
In contrast, conventional RC used reservoir layers with various numbers of neurons, and the hyperparameters were optimized.
Details of the models and their learning procedure are presented in Materials and Methods.

\subsection*{Simple attention tasks}

\begin{figure*}
\begin{center}
\includegraphics[clip,width=\textwidth]{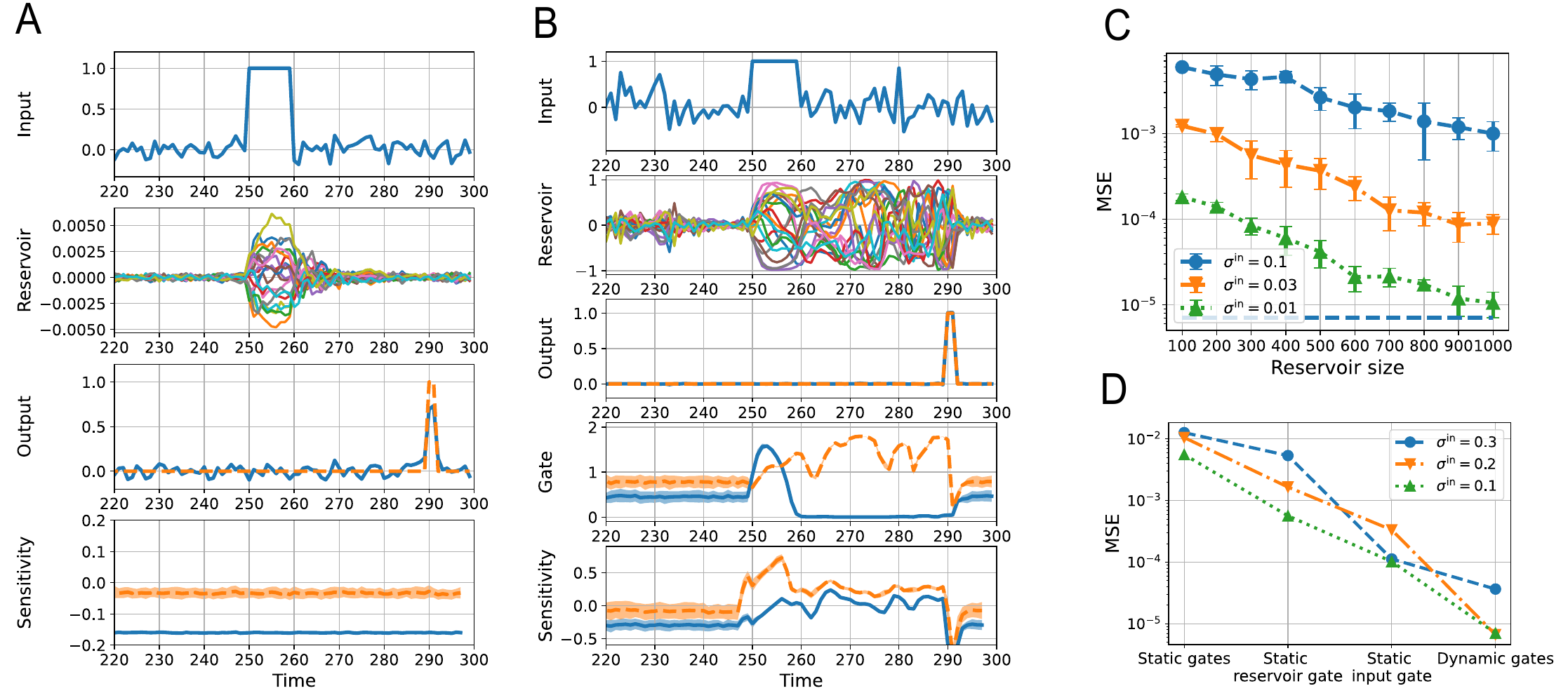}
\caption{Simulation results for simple attention tasks. A. Dynamics of a conventional RC model after training with $N^\text{res}=100$ and $\sigma ^\text{in}=0.1$.
From the top, the time evolution of the input signal, reservoir states, output, and sensitivity are shown. 
B. Dynamics of the SM-RC model after training with $N^\text{res}=100$ and $\sigma ^\text{in}=0.3$.
From the top, the time evolution of the input signal, reservoir states, outputs, gates, and sensitivity are shown. 
In the gate panel, the blue solid line represents the input gate, and the orange dashed line represents the spectral radius. 
The standard deviations of these gate values were obtained using 100 different input signals without jitter noise.
We show the modulated spectral radius $\rho ^\text{res}(t)$ obtained using Eq. \ref{eq:reservoir_gate}. 
In A and B, in the sensitivity panel, the blue solid line indicates the mean sensitivity of the reservoir layer, and the orange dashed line indicates the maximum sensitivity of the reservoir layer. 
The standard deviations of the sensitivity values were obtained using 100 different input signals without jitter noise.
For the output layer, the solid line indicates the predicted output, and the dashed line indicates the teacher signal. 
In the panels displaying the reservoir states, only 20 reservoir neurons are shown.
C. Comparison of regression errors (mean squared errors, MSEs) between conventional RC models and SM-RC models.
The regression errors for the conventional RC models are plotted as a function of the reservoir size for different input noise intensities ($\sigma ^\text{in}$). 
The standard deviations were obtained over 10 trials. 
The regression errors for the SM-RC models are represented by blue dashed horizontal lines for the input noise intensity of 0.1. 
For the SM-RC models, the reservoir size was fixed to 100. 
D. Regression errors for the SM-RC models when one or both gates are made static.
}
\label{fig:MemoryTask}
\end{center}
\end{figure*}

To investigate the self-modulation mechanism of SM-RC, we evaluated the learning performance for a simple attention task.
In this task, the input signal was $u(t)=1$ in the time interval of $[250,~259]$, and Gaussian noise with a mean of 0 and standard deviation of $\sigma ^\text{in}$ was added at other timesteps.
The model was trained to output $1$ in the time interval of $[290, 291]$ and output $0$ at other timesteps.
To cancel the effect of initial values of $x_i(0)=0$, the first 200 steps were discarded (free run), and common jitter noise was added to the input and output time windows (see Materials and Methods for details).
This task required a memory retention function that retained the input signal for a long time while reducing the influence of Gaussian noise when no informative input signals were provided.

Figure \ref{fig:MemoryTask} shows the learning results for the simple attention task.
In the case of conventional RC (Fig. \ref{fig:MemoryTask} A), for $\sigma ^\text{in}=0.1$, the regression of the output pulse was poor.
Although it has been reported that RC has a long-term memory function \cite{Jaeger2012long}, memory retention is difficult when noise is continuously added to the reservoir layer.
In contrast, for SM-RC (Fig. \ref{fig:MemoryTask} B), even when the noise was intensified, i.e., $\sigma ^\text{in}=0.3$, the regression of the output pulse was performed accurately.
Figure \ref{fig:MemoryTask} C presents a comparison of the regression performance between conventional RC and SM-RC.
SM-RC significantly outperformed conventional RC for various input noise intensities $\sigma ^\text{in}$.
For the same input noise intensity, SM-RC with $N^\text{res}=100$ outperformed conventional RC with a reservoir at least 10 times larger. 

By examining the time evolution of the input gate $g^\text{in}(t)$ and the reservoir gate $g^\text{res}(t)$ (the figure shows the modulated spectral radius $\rho ^\text{res}(t)$) (Fig. \ref{fig:MemoryTask} B), we can intuitively understand the factors contributing to successful regression in the case of SM-RC.
The input gate autonomously took large values during the time period when the input pulse arrives, efficiently feeding information into the reservoir layer.
After the input pulse disappears, the input gate takes small values to prevent the input noise from corrupting the information stored in the reservoir layer.
In contrast, the modulated spectral radius $\rho ^\text{res}(t)$ takes large values after the input pulse disappears. 
Because the fading memory condition is not met when the spectral radius is sufficiently large \cite{Dambre2012information, Lukosevicius2012practical}, 
the reservoir layer can retain the information for a long period.

To further study the dynamics of SM-RC, we performed a sensitivity analysis of the reservoir layer. 
This was done by adding perturbation to the reservoir states and examining the evolution of the perturbation two steps forward ($t_p=2$). 
The results for other values of $t_p$ are presented elsewhere (Appendix, Fig. \ref{fig:attention}). 
When the perturbation increases after $t_p$ steps, the sensitivity has a positive value; otherwise, it has a negative value (see Materials and Methods for details).
In Fig. \ref{fig:MemoryTask} A, the sensitivity of the conventional RC model exhibits only negative values during the simulation period.
This indicates that the reservoir was in stable states, which is consistent with the fading memory condition that usually holds for conventional RC \cite{Dambre2012information}. 
In contrast, for SM-RC (Fig. \ref{fig:MemoryTask} B), the sensitivity changed significantly over time and occasionally exhibited positive values. 
The positive sensitivity indicated that the reservoir was in chaotic states. 
Usually, chaotic states make regression difficult because the input--output relationship becomes sensitive to the initial conditions. 
However, in the case of SM-RC, by shutting the input gate, the problem related to the initial conditions was avoided for this attention task. 
We confirmed that the effects of Gaussian noise were negligible in the reservoir states after the input pulse and before the output pulse. 
The fact that the positive sensitivity only appeared after the input pulse and before the output pulse indicated that SM-RC successfully learned the complex reservoir dynamics adapted to the input signal.

To investigate the independent characteristics of the gates, we evaluated the regression performance when one or both gates did not evolve over time (see Materials and Methods for details).
Figure \ref{fig:MemoryTask} D shows the results.
For all input noise intensities, the performance was the best when both gates were time-evolved (dynamic gates), followed by the case where the reservoir gate was modulated (static input gate) and then the case where the input gate was modulated (static reservoir gate).
The performance was the worst when both gates were static (static gates).
This indicated that the observed performance improvement was due to not the optimization of the spectral radius and input intensity but their temporal modulation.
In addition, the fact that the performance was the best when both gates were time-modulated indicated that the input gate and reservoir gate were cooperatively modulated, as intuitively explained above.

\subsection*{Time-series prediction: NARMA and Lorentz model}

\begin{figure*}
\begin{center}
\includegraphics[clip,width=\textwidth]{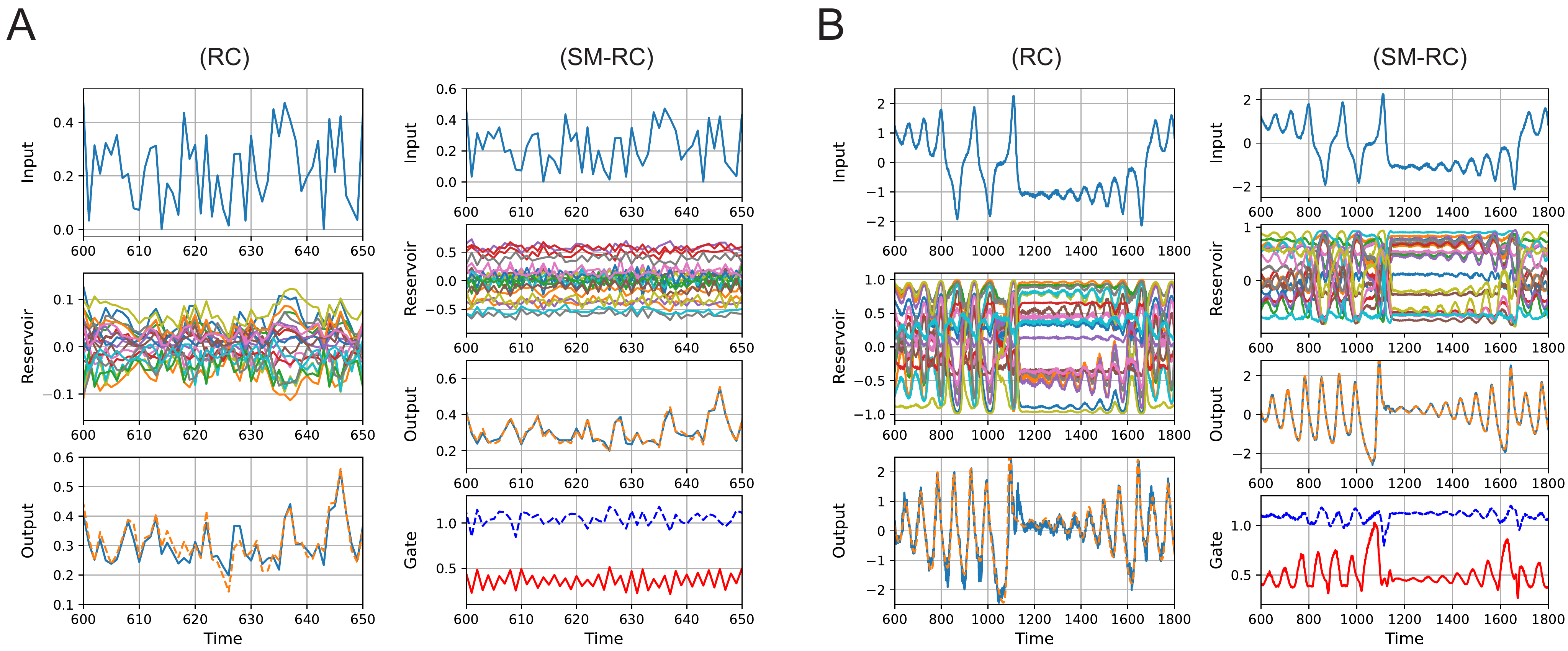}
\caption{Time evolution of the reservoir dynamics for (A) NARMA10 tasks with $N^\text{res}=100$ and (B) Lorentz model tasks with $N^\text{res}=100,~\sigma ^\text{in}=0.03$ and $N^\text{forward}=20$. In A and B, the left figures show the inputs, reservoir states, and outputs (from top to bottom) for the conventional RC models, and the right figures show the inputs, reservoir states, outputs, and gates (from top to bottom) for the SM-RC models. 
For the gates, the solid red lines represent the input gates, and the dashed blue lines indicate the modulated spectral radius. 
For the output layer, the solid line indicates the predicted output, and the dashed line indicates the teacher signal. 
In the panels displaying the reservoir states, only 20 reservoir neurons are shown.}
\label{fig:time_series}
\end{center}
\end{figure*}

SM-RC improves the prediction performance even when there is no apparently salient time series to which attention should be paid, as in the case of simple attention tasks.
To demonstrate this, we evaluated the performance of time-series prediction using NARMA and the Lorentz model.
The NARMA time series is obtained with a nonlinear autoregressive moving average, which is often used to evaluate the learning performance of RC \cite{Inubushi2017reservoir, Sakemi2020model, Jordanou2022investigation}. 
In particular, we use the NARMA5 and NARMA10 time series, which have different internal dynamics.
The Lorentz model is a chaotic dynamical system that is also used for RC performance evaluation owing to its difficulty of prediction \cite{Vlachas2020backpropagation, Chen2020autoreservoir, Lu2017reservoir, Katori2019reservoir, Inubushi2020transfer}. 
For the Lorentz model tasks, Gaussian noise with different standard deviations $\sigma^\text{in}$ was added to the input to approximate the situation of real-world data.
The aforementioned tasks involved input signals with different characteristics: a uniform random number for the NARMA tasks and a chaotic time series for the Lorentz model tasks (see Materials and Methods).

Figure \ref{fig:time_series} A shows the time evolution of the conventional RC and SM-RC models after they were trained on the NARMA10 task. 
Because the input signal for the task was uniform noise, the reservoir states evolved accordingly in the conventional RC and SM-RC models. 
In contrast to the case of the simple attention task, the gates in the SM-RC model did not change significantly over time, which reflected the characteristics of the input signal. 
The prediction performance of SM-RC was better than that of conventional RC, indicating that the self-modulation mechanism is effective even for uniform input signals.
Figure \ref{fig:time_series} B shows the time evolution of the conventional RC and SM-RC models after they were trained on the Lorentz model task. 
For both models, the time evolution of the reservoir states reflected the input chaotic time series.
Although it was not apparent in the time evolution of the reservoir states, we observed that the gates of the SM-RC model evolved adapting to the input chaotic time series.
The dynamics of SM-RC for various task conditions, i.e., $\sigma ^\text{in}$ and $N^\text{forward}$, are presented elsewhere (Appendix, Fig. \ref{fig:Lorentz_time_evolution}). 
Because of the dynamic behavior of the gates, the prediction performance of SM-RC was significantly better than that of conventional RC.

\begin{figure*}
\begin{center}
\includegraphics[clip,width=\textwidth]{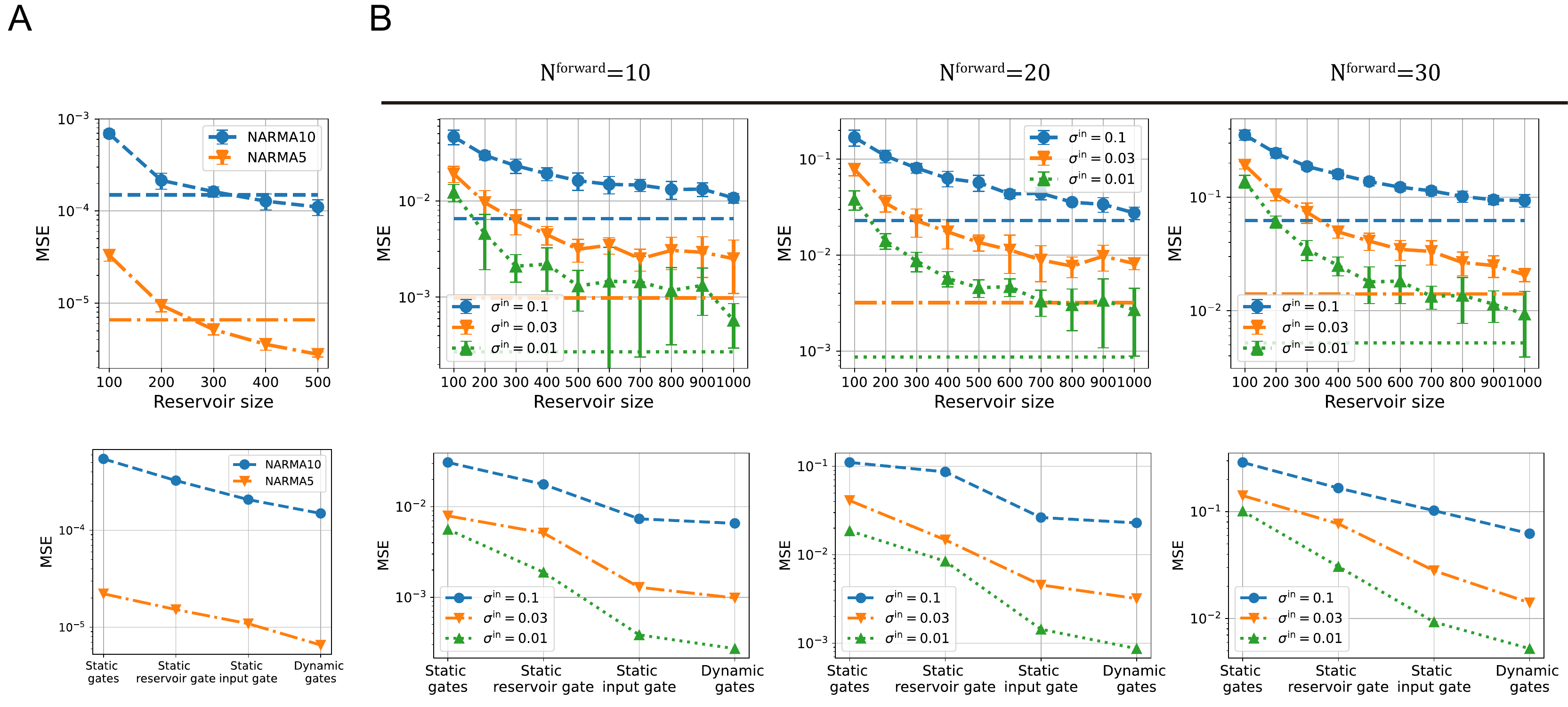}
\caption{Performance evaluation for the (A) NARMA tasks and (B) Lorentz model tasks. 
In A and B, the top figures show comparisons of the MSEs for the conventional RC and SM-RC models. 
In each figure, the MSEs for the conventional RC models are plotted. 
The horizontal axis indicates the size of the reservoir for the conventional RC model.
The standard deviations were obtained over 10 trials. 
The horizontal lines indicate the MSEs for the SM-RC models. 
The size of the SM-RC models was fixed to 100. 
The bottom figures show comparisons of the MSEs for the SM-RC models when some gates were temporally fixed. 
In B, the results for the cases of $N^\text{forward}=10$, $20$, and $30$ are presented in the left, center, and right figures, respectively. 
In addition, the results for different values of the input noise $\sigma ^\text{in}$ are presented. 
}
\label{fig:Lorentz_comparison}
\end{center}
\end{figure*}

Figure \ref{fig:Lorentz_comparison} shows a comparison of the prediction performance of conventional RC and SM-RC for NARMA10 (Fig. \ref{fig:Lorentz_comparison} A) and Lorentz model tasks (Fig. \ref{fig:Lorentz_comparison} B). 
In the top figures in Figs. \ref{fig:Lorentz_comparison} A and B, the prediction error of conventional RC is plotted as a function of the reservoir size. 
The horizontal lines indicate the SM-RC prediction error. The number of neurons in the reservoir layer of the SM-RC models were all fixed to 100.
For the Lorentz model tasks (Fig. \ref{fig:Lorentz_comparison} B), we present the results for different values of the following two parameters: the input noise $\sigma ^\text{in}$ and prediction steps $N^\text{forward}$.
As shown, SM-RC achieved significantly better prediction performance than conventional RC. 
For the NARMA tasks, SM-RC exhibited comparable prediction performance to conventional RC with a reservoir three times larger.
In contrast, for the Lorentz model tasks, SM-RC exhibited prediction performance comparable to conventional RC with a reservoir 10 times larger.

In the bottom figures in Figs. \ref{fig:Lorentz_comparison} A and B, the prediction performance of SM-RC when one or both gates do not evolve over time is shown.
For the NARMA and Lorentz model tasks, the performance was the best when both gates evolved, followed by the case where the reservoir gate was modulated and then the case where the input gate was modulated.
The performance was the worst when both gates were static.
These results are similar to those for the simple attention task. 
As before, they indicate that the observed performance improvement was due to not the optimization of the spectral radius and input intensity but their temporal modulation.

\begin{figure}
\begin{center}
\includegraphics[clip,width=8cm]{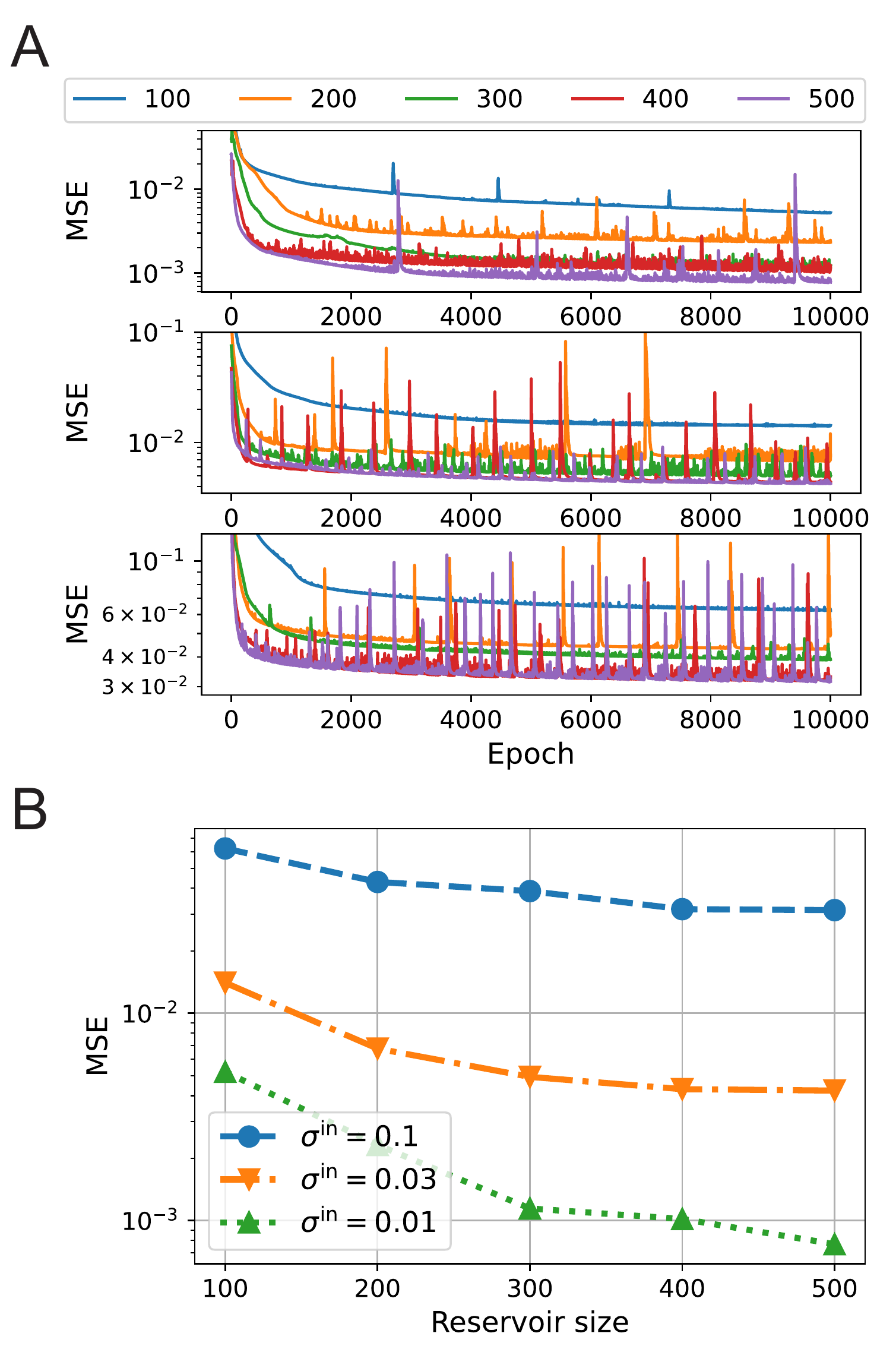}
\caption{Regression performance of SM-RC models with different reservoir sizes for the Lorentz model tasks ($N^\text{forward}=30$).
A. MSEs at each training step (epoch) for $\sigma ^\text{in}=0.01$, 0.03, and 0.1, from top to bottom.
B. MSEs for SM-RC models with different reservoir sizes.} 
\label{fig:Lorentz_size}
\end{center}
\end{figure}

Finally, to investigate the scalability of SM-RC, we evaluated the prediction performance for the Lorentz model task with an increasing reservoir size. 
Figure \ref{fig:Lorentz_size} A shows the learning curves (MSEs with respect to the learning epoch). 
From top to bottom, the panels show the results for SM-RC with different reservoir sizes ($N^\text{size}=100,~200,\dots,500$) when the input noise was 0.01, 0.03, and 0.1. 
The learning curve became spiky as the reservoir size increased, possibly because of the difficulties in RNN learning (Appendix, Fig. \ref{fig:learning_curves}). 
Nevertheless, the training of SM-RC was successful. 
Figure \ref{fig:Lorentz_size} B shows the resulting performance of SM-RC for different reservoir sizes. 
The MSEs decreased monotonically as the reservoir size increased, indicating that SM-RC is scalable with regard to the learning performance.

\section*{Discussion}

We propose SM-RC, which can alter the dynamical properties of the reservoir layer by introducing gate structures. 
We found that the SM-RC architecture achieved better learning performance than conventional RC for simple attention, NARMA, and Lorentz model tasks. 
In particular, the improvement in the learning performance for the simple attention task and the Lorentz model task was remarkable. 
In the case of non-uniform input signals, the mechanism that adapt the way of information processing to the input signal is considered to be effective.
Because most real-world time-series data are non-uniform, the proposed architecture is expected to be effective for real-world data.
In the simple attention task, we observed local chaotic dynamics in SM-RC. 
Chaotic dynamics are of interest in deep-learning research because they increase the expressiveness of learning models \cite{Poole2016exponential, Zhang2020expressivity}. 
Recently, Inoue {\it et al.} observed transient chaos in transformer models \cite{Inoue2022transient}.
It is expected that SM-RC utilized the high expressivity of chaotic dynamics for encoding the input signal and outputting the pulse. 
However, the underlying learning mechanism requires further investigation.
We expect that detailed analysis of dynamical systems \cite{Dambre2012information, Barak2013from, Sussilo2013opening, Rivkind2017local} will produce ideas for improving SM-RC.

SM-RC provides insights into the mechanism of the nervous system.
Neural networks with random connections, as used in RC, are attracting attention as models of the brain \cite{Maass2002realtime,Barak2013from, Seoane2019evolutionary, Suarez2021learning}. 
The proposed SM-RC model can be thought of as incorporating the mechanism of neuromodulation into RC.
This is because both systems globally and temporally modulate the activity levels of neurons \cite{Thiele2018neuromodulation, Edelmann2018dopaminergic, Palancios2019neuro}.
For example, the neuromodulator acetylcholine is delivered from the basal forebrain to brain areas, increasing the neuronal activity therein \cite{Thiele2018neuromodulation}.
SM-RC realizes an attention mechanism that can be related to neuromodulation \cite{Thiele2018neuromodulation}.
We expect that the resemblance between SM-RC and the neuromodulation mechanism can be studied by adopting biologically plausibility characteristics such as spike-based communications, 
synaptic weights and time constants, and neural geometries. 

Efficient hardware implementation is important for real-world implementation of SM-RC \cite{Tanaka2019recent,Nakajima2020physical}.
Because the input gate $g^\text{in}$ only directly modulates the input intensity, it is not difficult to implement.
Additionally, we believe that the reservoir gate $g^\text{res}$ can be implemented according to the hardware mechanism.
For example, in analog circuit implementations, connections between neurons are represented by current magnitudes \cite{Bauer2019real,Yamaguchi2019chaotic}.
If the connection weights are implemented with digital memory, it is sufficient to provide a mechanism for modulating the digital values.
In the case of analog memory, if it is implemented with three-terminal memory devices such as floating-gate devices, by modulating the gate voltage of all or part of the field-effect transistor of the memory, modulation of reservoir-layer dynamics can be achieved.
In addition, delay feedback systems \cite{Appeltant2011information} can build a reservoir layer using a single nonlinear node, which is often used in photonic systems \cite{Brunner2018tutorial} and electronic integrated circuits \cite{Penkovsky2018efficient}. 
In these systems, the entire reservoir system can be modulated simply by changing the single node; thus, it is relatively easy to implement the reservoir gate.
This hardware-friendly aspect of SM-RC is what motivates us to use SM-RC for edge AI rather than other state-of-the RNN models such as LSTM \cite{Hochreiter1997long} and GRUs \cite{Cho2014learning}. 

We believe that it is possible to train SM-RC implemented in physical systems using various learning techniques.
In this paper, we trained the SM-RC model via backpropagation through time (BPTT) \cite{Werbos1990backpropagation}. 
In BPTT, detailed information regarding the reservoir-layer dynamics is required, but in physical and analog systems, it is often difficult to capture the internal dynamics in detail.
Efforts to train such hardware have progressed in recent years.
For example, Wright et al. successfully trained a black-boxed system by capturing the internal dynamics using deep learning \cite{Wright2022deep}. 
In addition, error backpropagation approximations, such as feedback alignment \cite{Lillicrap2016random} and its variants \cite{Nokland2016direct, Guerguiev2017towards, Murray2019local, Frenkel2021learning}, are suitable for training models implemented in physical systems. 
It has been reported that BPTT can be approximated so that RNNs can be trained in a biologically plausible manner \cite{Murray2019local, Bellec2020solution}.
Nakajima et al. extended these methods and successfully trained multilayer physical RC \cite{Nakajima2022physical}.
By applying those methods to SM-RC, it is expected that SM-RC can be trained in analog and physical systems. 
We believe that SM-RC represents a new paradigm for physical RC.

\section*{Materials and Methods}

\subsection*{Model}

The time evolution of the ESN-type SM-RC model can be expressed as
\begin{flalign}
\bm{x}(t) &= \tanh \bigl( g^\text{res}(t-1) W^\text{res} \bm{x}(t-1) \nonumber\\
&~~ + g^\text{in}(t-1) W^\text{in} \bm{u}(t) + \xi \mathbf{1} \bigr), \label{eq:SM_RC}\\
g^\text{res} (t) &= f \left( W^\text{res}_\text{fb} \bm{x}(t) + b^\text{res}_\text{fb} \right), \label{eq:g_res}\\
g^\text{in} (t) &= f \left( W^\text{in}_\text{fb} \bm{x}(t) + b^\text{in}_\text{fb} \right), \label{eq:g_in}\\
y(t) &= W^\text{out} \bm{x}(t) + b^\text{out},
\end{flalign}
where $\bm{x}(t) \in \mathbb{R}^{N^\text{res}}$ and $y(t) \in \mathbb{R}$ represent the internal state and output of the reservoir layer, respectively.
$W^\text{in} \in \mathbb{R}^{N^\text{res}\times N^\text{in}}$ and $W^\text{res} \in \mathbb{R}^{N^\text{res}\times N^\text{res}}$ are matrices representing the connection weights from the input to the reservoir and the inner connection weights of the reservoir, respectively. 
$N^\text{in}$ and $N^\text{res}$ represent the input dimension and reservoir dimension, respectively.
Each element of $W^\text{in}$ is initialized by randomly sampling from the uniform distribution of $[-\rho^\text{in}, \rho^\text{in}]$.
Each element of $W^\text{res}$ is randomly sampled from the uniform distribution of $[-1,1]$; then, it is multiplied a constant so that the ``initial'' spectral radius becomes $\hat{\rho} ^\text{res}$.
$\mathbf{1}$ is a vector of 1s, and $\xi \in \mathbb{R}$ controls the magnitude of the bias term of neurons in the reservoir layer \cite{Lu2017reservoir}.
$g^\text{in}(t) \in \mathbb{R}$ is the input gate that modulates the input intensity and is controlled by feedback from the reservoir layer via weights $W_\text{fb}^\text{in}$.
Similarly, $g^\text{res}(t) \in \mathbb{R}$ is the reservoir gate that modulates the connection strength of the reservoir and is controlled by feedback from the reservoir layer via weights $W_\text{fb}^\text{res}$.
The output function $f$ of the gate is a non-negative function. In this study, the output value was limited to $(0,2)$ to stabilize the learning process: 
\begin{flalign}
f(x) &= \frac{2}{1+e^{-x}}. 
\end{flalign}
The input gate temporally modulates the intensity of the input by multiplying it by a value within the range of (0, 2), allowing the selection of useful input signals. 
In contrast, the reservoir gate temporally modulates the spectral radius $\rho ^\text{res}$ within $(0, 2\hat{\rho} ^\text{res})$, allowing the information stored in the reservoir layer to be retained or discarded. 
The time evolution of conventional RC is obtained by replacing both $g^\text{in}$ and $g^\text{res}$ with 1. 

$W^\text{in}$ and $W^\text{res}$ are fixed weights, as in the conventional RC framework. 
In SM-RC, in addition to the output weights ($W^\text{out}$), feedback weights ($W_\text{fb}^\text{in}, b_\text{fb}^\text{res}, W_\text{fb}^\text{res}, b_\text{fb}^\text{res}$) are trained. 
In this study, the output weights were trained using the pseudo-inverse method, and the feedback weights were trained using BPTT \cite{Werbos1990backpropagation}) (see below).

\subsection*{Learning procedure}

All the tasks performed in this study involved predicting the target signal $y^\text{t}(t)$ using the input time series \{${u(0), u(1), \dots u(t)}$\} obtained by time $t$.
RC usually inputs only $u(t)$ to the reservoir layer at time $t$ (Eq. \ref{eq:SM_RC}). 
In the case of conventional RC, training of the output weights was performed via the pseudo-inverse method to minimize the squared error between the output and the target signal \cite{Lukosevicius2012practical}:
\begin{flalign}
\sum _{t=1}^T \|y(t) - y^\text{t}(t) \|^2, \label{eq:LossFunction}
\end{flalign}
where $T$ represents the number of timesteps of the training data.
When there are multiple training data, $T$ is the product of the number of timesteps of the training data and the number of training data.
To avoid the influence of the initial state, the first 200 steps in the simulation were excluded from the training data (free run).
In this study, adding the regularization term $\|W^\text{out}\|$ (ridge regression) did not improve the performance.

In the case of SM-RC, we used the same loss function (Eq. \ref{eq:LossFunction}) and trained weights via the gradient descent method.
In each epoch, the weights were updated via full-batch training, and this was repeated for $10^4$ epochs.
In each full-batch training process, the output layer was trained using the pseudo-inverse method as in conventional RC, and then the gates were trained using BPTT.
In BPTT, the weights were updated using the Adam optimizer with a learning rate of $10^{-3}$ \cite{Kingma2014adam}.
All implementations and simulations were performed using the PyTorch framework \cite{Pazke2019pytorch}.
It is also possible to uniformly train all the learning parameters, including the output weights, via BPTT. However, in this case, the convergence of the learning was slow and the learning process became unstable, degrading the learning performance. 
By learning the output layer in advance, the backpropagating error signal is reduced, which may lead to stable learning (Appendix, Fig. \ref{fig:learning_curves}).

The hyperparameters of the conventional RC model ($\rho ^\text{in},~\rho ^\text{res}$, and $\xi$) were optimized via Bayesian optimization \cite{Snoek2012practival,Frazier2018tutorial}.
We set $\xi=0$ for the simple attention tasks and NARMA tasks.
The hyperparameters of the SM-RC model were set as $\rho ^\text{in}=0.12,~\hat{\rho} ^\text{res}=0.9$. The bias term was set as $\xi=0$ for the simple attention tasks and NARMA tasks and $\xi=0.2$ for the Lorentz model tasks.
We observed that SM-RC training was somewhat unstable (Appendix, Fig. \ref{fig:learning_curves}).
Therefore, in the performance evaluation of SM-RC, we considered the best results for training with 50 different initial weights.

In this study, we investigated the case where a single gate or both gates did not change over time.
This was done by fixing all the weight elements ($W_\text{fb}^\text{in (res)}$) to $0$ and training only the bias term $b_\text{fb}^\text{in (res)}$.

\subsection*{Datasets}

In the simple attention task, the $i$th input training datum had a value of 1 in the input time interval $[250 + t_i^\text{jitter},~ 259 + t_i^\text{jitter}]$ and a value sampled from a Gaussian distribution with a mean of 0 and a standard deviation of $\sigma ^\text{in} \in \{0.1, 0.2, 0.3\}$ at other timesteps.
The $i$th output training datum had a value of $1$ in the time interval of $[290 + t_i^\text{jitter}, 291 + t_i^\text{jitter}]$ and $0$ at other timesteps.
$t_i^\text{jitter}$ represents the jitter noise; one of $\{-2,-1, 0, 1, 2\} \in Z$ was randomly sampled uniformly.
The jitter noise was introduced to prevent the learning models from generating output pulses using the initial value ($x_i(0) = 0$) instead of using the input signal.
We used 100 data obtained as described above as training data and 100 other data as test data.

NARMA is a nonlinear autoregressive moving average model given by
\begin{flalign}
&y^\text{tc}(t) = ~0.3y^\text{tc}(t-1) + 0.05y^\text{tc}(t-1) \sum _{i=1} ^{m} y^\text{tc}(t-i) + 1.5s(t-m+1) s(t) + 0.1, 
\end{flalign}
where $s(t)\in \mathbb{R}$ is uniformly sampled from the interval $[0, 0.5]$.
The NARMA5 and NARMA10 time series were obtained with $m = 5$ and $m = 10$, respectively. 
The task was to predict $y^\text{tc}(t)$ using $s(t)$ as the input.
To eliminate the influence of the initial value, the first 200 steps were discarded, and 2000 steps were generated as training data.
We also generated test data for 2000 steps using the same method.
We only used a single data for the training dataset and a single data for the test dataset.

The Lorentz model evolves over time according to the following differential equation:
\begin{flalign}
\frac{dx}{dt}=10 (y-x), \frac{dy}{dt}=x(28 - z) -y, \frac{dz}{dt} = xy - \frac{8}{3}z.
\end{flalign}
Discrete time-series data were obtained from the above differential equation by using the Euler method with a timestep of $\Delta t = 0.01$.
To eliminate the influence of the initial value, the first $1000$ steps were discarded, and the time series of the subsequent $2000$ steps was obtained.
100 training data and 100 different test data were generated in the same way from different initial values.
The task was to predict $z(t+N_\text{forward}\Delta t)$ with $x(t)$ as the input.
$N_\text{forward}\in \{10, 20, 30\}$ represents the number of steps forward to predict.
$x(t)$ and $z(t)$ were normalized to a mean of $0$ and variance of $1$ for the training and test data, respectively. 
We also added Gaussian noise with a mean of 0 and variance of $\sigma ^\text{in} \in \{0.01, 0.03, 0.1\}$ to the input $x(t)$ in the training data.

\subsection*{Sensitivity analysis}

We evaluated the sensitivity of the learning models in the simple attention task.
The sensitivity indicates how a perturbation to the reservoir state is magnified in $t_p$ timesteps. 
For the input data without jitter noise, SM-RC was operated, and the states of the reservoir layer are taken as the reference states $\{\bm{x}^\text{base}(0), \bm{x}^\text{base}(1), \dots, \bm{x}^\text{base}(t), \dots \}$.
Then, a perturbation $p\in \mathbb{R}^{N^\text{res}}$ is applied to the reservoir state $\bm{x}^\text{base}(t)\in \mathbb{R}^{N^\text{res}}$ at time $t$, and the reservoir state $\bm{x}^{p_j}(t+t_p)\in \mathbb{R}^{N^\text{res}}$ at time $t + t_p$ is obtained using Eq. \ref{eq:SM_RC}.
The perturbation vector was obtained by sampling a vector contained in the unit circle using the Metropolis--Hastings algorithm, followed by rescaling so that the obtained vector satisfied $\|p_j\|=\epsilon$.
We estimated the sensitivity using the reference and perturbed states as follows:
\begin{flalign}
\lambda (t) = \frac{1}{t_p N_p} \sum _{j=1}^{N_p} \ln \left(\frac{\|\bm{x}^\text{base}(t+t_p) - \bm{x}^{p_j}(t+t_p)\|}{\epsilon}\right),
\end{flalign}
where $N_p$ represents the number of perturbation vectors. 
Similarly, we estimated the maximum sensitivity:
\begin{flalign}
\lambda _\text{max}(t) = \frac{1}{t_p}\text{max}_j\left[\ln \left(\frac{\|\bm{x}^\text{base}(t+t_p) - \bm{x}^{p_j}(t+t_p)\|}{\epsilon}\right)\right].
\end{flalign}
In the simulation, we used $t_p =2$, $\epsilon = 10^{-8}$, and $N_p=200$. 
The simulation results when other values of $t_p$ were used are presented elsewhere (Appendix, Fig. \ref{fig:attention}).
The perturbation to the gates $g^\text{res},~g^\text{in}$ was applied indirectly in accordance with Eq. \ref{eq:g_res} because they are not independent (state) variables. 
The gates $g^\text{res}, g^\text{in}$ are dependent variables because they are determined by the state of the reservoir layer.
Therefore, in the case of SM-RC, the perturbation is applied only to the state of the neurons in the reservoir layer.
 Note that the maximum sensitivity is also known as the maximum local Lyapunov exponent. 

\section*{Acknowledgements}
This work was partially supported by SECOM Science and Technology Foundation, 
JST PRESTO Grant Number JPMJPR22C5, 
JST Moonshot R\&D Grant Number JPMJMS2021, 
AMED under Grant Number JP22dm0307009, 
Institute of AI and Beyond of UTokyo, 
the International Research Center for Neurointelligence (WPI-IRCN) at The University of Tokyo Institutes for Advanced Study (UTIAS), 
JSPS KAKENHI Grant Number JP20H05921. 
Computational resource of AI Bridging Cloud Infrastructure (ABCI) provided by National Institute of Advanced Industrial Science and Technology (AIST) was used.

\appendix

\section{Appendices}

\subsection{Dependence of timesteps $t_p$ in simple attention task}

\begin{figure*}
\begin{center}
\includegraphics[clip,width=\textwidth]{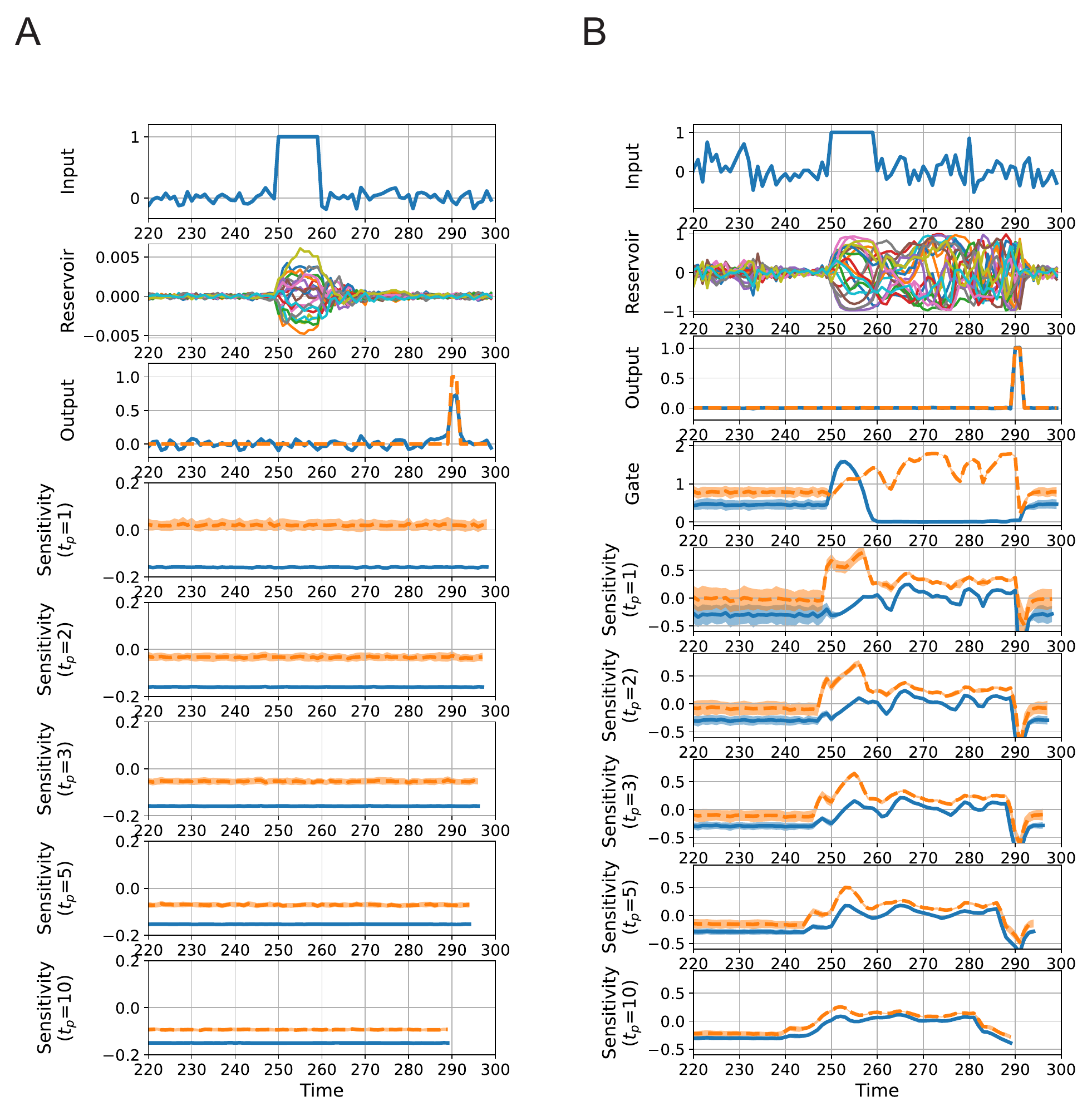}
\caption{Results of the sensitivity analysis for the simple attention task. 
A. Results for the case of RC with $N^\text{res}=100$ and $\sigma ^\text{in}=0.1$. 
From the top, the time evolution of the input signal, reservoir states, output, and sensitivity are shown. 
B. Results for the case of SM-RC with $N^\text{res}=100$ and $\sigma ^\text{in}=0.3$.
From the top, the time evolution of the input signal, reservoir states, outputs, gates, and sensitivity are shown.
In A and B, the sensitivity is analyzed using different timesteps $t_p$, at which the magnitudes of expansion of the perturbations were measured. 
For the output layer, the solid line indicates the predicted output, and the dashed line indicates the teacher signal. 
In the panels displaying the reservoir states, only 20 reservoir neurons are shown.}
\label{fig:attention}
\end{center}
\end{figure*}

We performed the sensitivity analysis for the simple attention task using different timesteps $t_p$, at which the magnitudes of expansion of the perturbations were measured. 
Figure \ref{fig:attention} shows the results for the RC and SM-RC models. 

\subsection{Learning stability}

\begin{figure*}
\begin{center}
\includegraphics[clip,width=\textwidth]{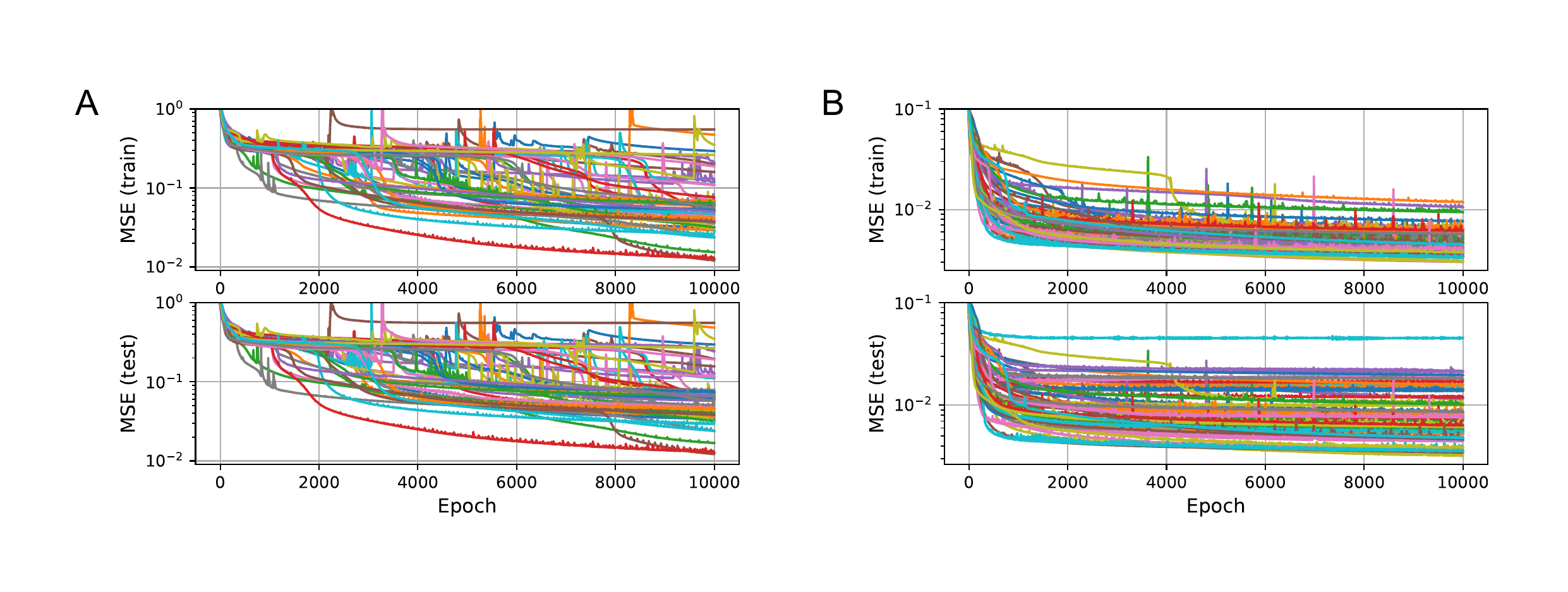}
\caption{MSEs as a function of the learning epoch in Lorentz model tasks with $N_\text{forward}=20,~\sigma=0.03$. 
50 different trials are shown with different initial weights for cases where (A) the output layer was also trained via BPTT and (B) the output layer was trained via the pseudo-inverse method. }
\label{fig:learning_curves}
\end{center}
\end{figure*}

Recurrent neural network (RNN) learning is unstable because there are many regions where the loss function changes the value sharply \cite{Doya1992bifurcations,Pascanu2013on}.
Reservoir computing (RC) avoids this problem by learning only the output layer \cite{Jaeger2001echo}.
In self-modulated RC (SM-RC), we found that the learning process was unstable, which is attributed to the fact that the RNN was indirectly trained through gates.
Figure \ref{fig:learning_curves} shows the MSE for each learning epoch in 50 different trials.
The learning task involved the Lorentz model, with $N^\text{forward}=20$ and $\sigma^\text{Lorentz}=0.03$.
Figure \ref{fig:learning_curves} A shows the case where all the trainable weights, including the output layer, were trained via the gradient descent method.
In this case, the learning process was unstable, resulting in poor performance.
This implies that SM-RC training has the same problems as RNN training described above.
Figure \ref{fig:learning_curves} B shows the case where the weights of the output layer were trained via the pseudo-inverse method and the weights related to the gates were learned via the gradient method.
In this case, the learning process was stabilized, and the performance was significantly improved. 
By training the output layer in advance, the loss is reduced; consequently, the backpropagating error signal is suppressed, which is thought to lead to stabilization.
However, because the resulting performance still varied among trials, we evaluated the learning performance of SM-RC by repeating the training procedure with different initial weights.
Gradient clip \cite{Pascanu2013on}, which is known as a method to stabilize learning, did not improve the performance

\subsection{Time evolution of SM-RC for Lorentz model tasks}

\begin{figure*}
\begin{center}
\includegraphics[clip,width=\textwidth]{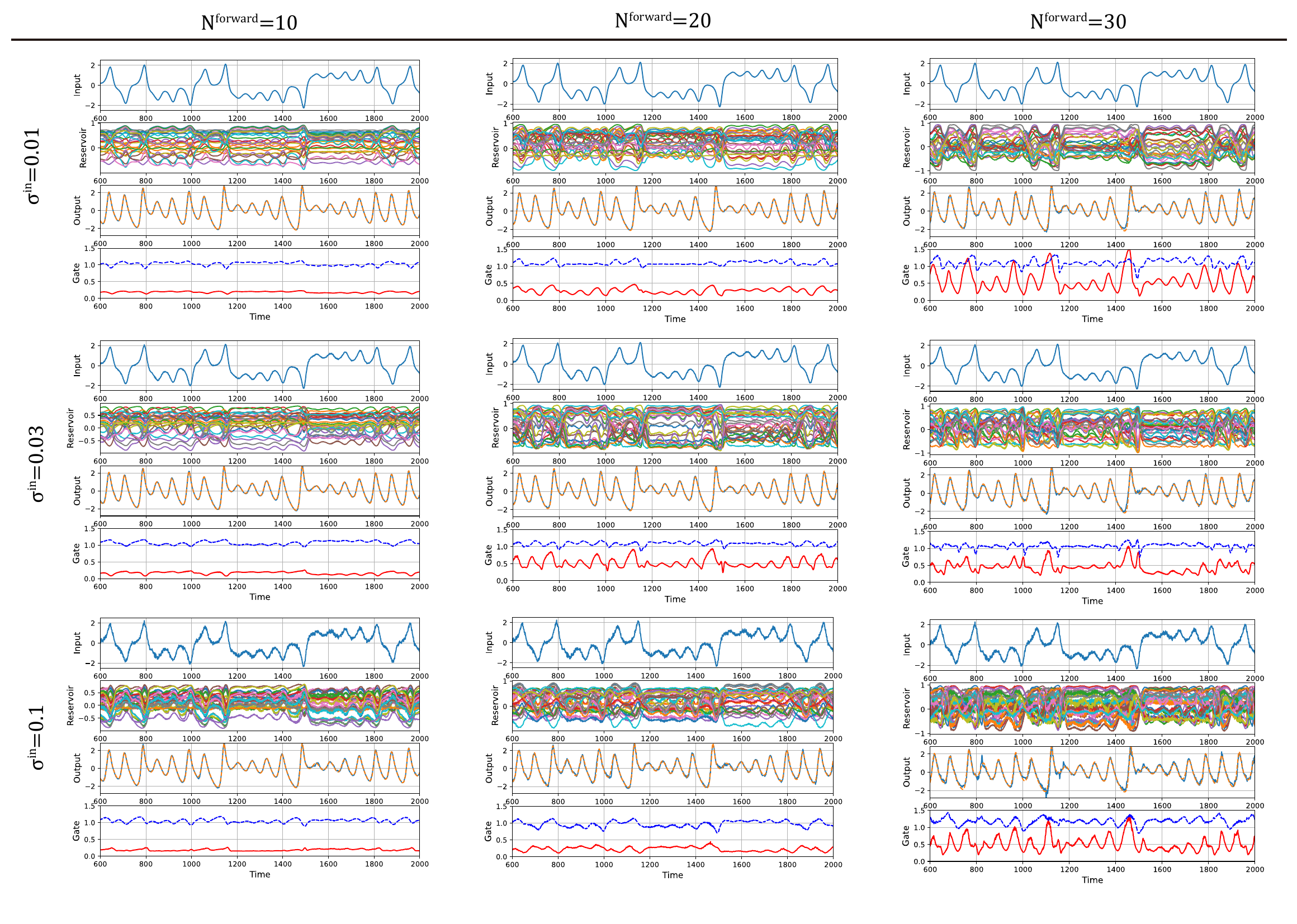}
\caption{Time evolution of SM-RC for the Lorentz model tasks with various input noises $\sigma ^\text{in}$ and numbers of prediction steps $N^\text{forward}$.
For the output layer, the solid lines indicate the predicted outputs, and the dashed lines indicate the teacher signals. 
For the gates, the solid red lines indicate the input gates, and the blue dashed lines indicate the spectral radius.
In the panels displaying the reservoir states, only 20 reservoir neurons are shown.
}
\label{fig:Lorentz_time_evolution}
\end{center}
\end{figure*}

Figure \ref{fig:Lorentz_time_evolution} shows the time evolution of the SM-RC models trained on the Lorentz tasks when the task conditions $\sigma^\text{in},~N^\text{forward}$ were changed.
As shown, in all the cases, the gates were modulated to adapt to the input signal.
The behaviors of the gates were complex and depended on the task conditions, 
indicating that the self-modulation mechanism can adapt the reservoir dynamics to tasks in a flexible manner.

\bibliographystyle{unsrt}
\bibliography{myBib}

\begin{thebibliography}{10}

\bibitem{Ismail2019deep}
Hassan Ismail~Fawaz, Germain Forestier, Jonathan Weber, Lhassane Idoumghar, and
  Pierre-Alain Muller.
\newblock Deep learning for time series classification: a review.
\newblock {\em Data mining and knowledge discovery}, 33(4):917--963, 2019.

\bibitem{Dong2021survey}
Shi Dong, Ping Wang, and Khushnood Abbas.
\newblock A survey on deep learning and its applications.
\newblock {\em Computer Science Review}, 40:100379, 2021.

\bibitem{Thompson2020computational}
Neil~C. Thompson, Kristjan Greenewald, Keeheon Lee, and Gabriel~F. Manso.
\newblock The computational limits of deep learning.
\newblock {\em arXiv}, 2007.05558, 2020.

\bibitem{Patterson2021carbon}
David Patterson, Joseph Gonzalez, Quoc Le, Chen Liang, Lluis-Miquel Munguia,
  Daniel Rothchild, David So, Maud Texier, and Jeff Dean.
\newblock Carbon emissions and large neural network training.
\newblock {\em arXiv}, 2104.10350, 2021.

\bibitem{Murshed2021machine}
M.~G.~Sarwar Murshed, Christopher Murphy, Daqing Hou, Nazar Khan, Ganesh
  Ananthanarayanan, and Faraz Hussain.
\newblock Machine learning at the network edge: A survey.
\newblock {\em ACM Comput. Surv.}, 54(8):1--37, oct 2021.

\bibitem{Jaeger2001echo}
H.~Jaeger.
\newblock The “echo state” approach to analysing and training recurrent
  neural networks.
\newblock {\em Technical Report GMD Report 148, German National Research Center
  for Information Technology}, 2001.

\bibitem{Maass2002realtime}
W.~Maass, T.~Natschläger, and H.~Markram.
\newblock Real-time computing without stable states: A new framework for neural
  computation based on perturbations.
\newblock {\em Neural Computation}, 14(11):2531--2560, 2002.

\bibitem{Hochreiter1997long}
Sepp Hochreiter and Jürgen Schmidhuber.
\newblock Long short-term memory.
\newblock {\em Neural Computation}, 9(8):1735--1780, 1997.

\bibitem{Cho2014learning}
Kyunghyun Cho, Bart van Merri{\"e}nboer, Caglar Gulcehre, Dzmitry Bahdanau,
  Fethi Bougares, Holger Schwenk, and Yoshua Bengio.
\newblock Learning phrase representations using {RNN} encoder{--}decoder for
  statistical machine translation.
\newblock In {\em Proceedings of the 2014 Conference on Empirical Methods in
  Natural Language Processing ({EMNLP})}, pages 1724--1734, Doha, Qatar,
  October 2014.

\bibitem{Dambre2012information}
J.~Dambre, D.~Verstraeten, B.~Schrauwen, and S.~Massar.
\newblock Information processing capacity of dynamical systems.
\newblock {\em Scientific reports}, 2:514, 2012.

\bibitem{Tanaka2019recent}
Gouhei Tanaka, Toshiyuki Yamane, Jean~Benoit Héroux, Ryosho Nakane, Naoki
  Kanazawa, Seiji Takeda, Hidetoshi Numata, Daiju Nakano, and Akira Hirose.
\newblock Recent advances in physical reservoir computing: A review.
\newblock {\em Neural Networks}, 115:100--123, 2019.

\bibitem{Nakajima2020physical}
Kohei Nakajima.
\newblock Physical reservoir computing{\textemdash}an introductory perspective.
\newblock {\em Japanese Journal of Applied Physics}, 59(6):060501, may 2020.

\bibitem{Vlachas2020backpropagation}
P.R. Vlachas, J.~Pathak, B.R. Hunt, T.P. Sapsis, M.~Girvan, E.~Ott, and
  P.~Koumoutsakos.
\newblock Backpropagation algorithms and reservoir computing in recurrent
  neural networks for the forecasting of complex spatiotemporal dynamics.
\newblock {\em Neural Networks}, 126:191--217, 2020.

\bibitem{Sum2020review}
Chenxi Sun, Moxian Song, Shenda Hong, and Hongyan Li.
\newblock A review of designs and applications of echo state networks.
\newblock {\em arXiv}, 2012.02974, 2020.

\bibitem{Tong2018reservoir}
Zhiqiang Tong and Gouhei Tanaka.
\newblock Reservoir computing with untrained convolutional neural networks for
  image recognition.
\newblock In {\em 2018 24th International Conference on Pattern Recognition
  (ICPR)}, pages 1289--1294, 2018.

\bibitem{Pathak2018model}
Jaideep Pathak, Brian Hunt, Michelle Girvan, Zhixin Lu, and Edward Ott.
\newblock Model-free prediction of large spatiotemporally chaotic systems from
  data: A reservoir computing approach.
\newblock {\em Phys. Rev. Lett.}, 120:024102, Jan 2018.

\bibitem{Kawai2022self}
Yuji Kawai, Jihoon Park, Ichiro Tsuda, and Minoru Asada.
\newblock Self-organization of a dynamical orthogonal basis acquiring large
  memory capacity in modular reservoir computing.
\newblock In {\em International Conference on Artificial Neural Networks},
  pages 635--646, 2022.

\bibitem{Gallicchio2017deep}
Claudio Gallicchio, Alessio Micheli, and Luca Pedrelli.
\newblock Deep reservoir computing: A critical experimental analysis.
\newblock {\em Neurocomputing}, 268:87--99, 2017.

\bibitem{Sakemi2020model}
Yusuke Sakemi, Kai Morino, Timoth{\'e}e Leleu, and Kazuyuki Aihara.
\newblock Model-size reduction for reservoir computing by concatenating
  internal states through time.
\newblock {\em Scientific reports}, 10:21794, 2020.

\bibitem{Chen2020autoreservoir}
Pei Chen, Rui Liu, Kazuyuki Aihara, and Luonan Chen.
\newblock Autoreservoir computing for multistep ahead prediction based on the
  spatiotemporal information transformation.
\newblock {\em Nature communications}, 11(1):4568, 2020.

\bibitem{Sussilo2009generating}
David Sussillo and L.F. Abbott.
\newblock Generating coherent patterns of activity from chaotic neural
  networks.
\newblock {\em Neuron}, 63(4):544--557, 2009.

\bibitem{Rivkind2017local}
Alexander Rivkind and Omri Barak.
\newblock Local dynamics in trained recurrent neural networks.
\newblock {\em Phys. Rev. Lett.}, 118:258101, Jun 2017.

\bibitem{Nicola2017supervised}
Wilten Nicola and Claudia Clopath.
\newblock Supervised learning in spiking neural networks with force training.
\newblock {\em Nature communications}, 8(1):2208, 2017.

\bibitem{Maxtsuki2019adaptive}
Toshitaka Matsuki and Katsunari Shibata.
\newblock Adaptive balancing of exploration and exploitation around the edge of
  chaos in internal-chaos-based learning.
\newblock {\em Neural Networks}, 132:19--29, 2020.

\bibitem{Schrauwen2008impriving}
Benjamin Schrauwen, Marion Wardermann, David Verstraeten, Jochen~J. Steil, and
  Dirk Stroobandt.
\newblock Improving reservoirs using intrinsic plasticity.
\newblock {\em Neurocomputing}, 71(7):1159--1171, 2008.

\bibitem{Yusoff2016modeling}
Mohd-Hanif Yusoff, Joseph Chrol-Cannon, and Yaochu Jin.
\newblock Modeling neural plasticity in echo state networks for classification
  and regression.
\newblock {\em Information Sciences}, 364-365:184--196, 2016.

\bibitem{Morales2021unveiling}
Guillermo~B. Morales, Claudio~R. Mirasso, and Miguel~C. Soriano.
\newblock Unveiling the role of plasticity rules in reservoir computing.
\newblock {\em Neurocomputing}, 461:705--715, 2021.

\bibitem{Laje2013robust}
Rodrigo Laje and Dean~V Buonomano.
\newblock Robust timing and motor patterns by taming chaos in recurrent neural
  networks.
\newblock {\em Nature neuroscience}, 16(7):925--933, 2013.

\bibitem{Inoue2020designing}
Katsuma Inoue, Kohei Nakajima, and Yasuo Kuniyoshi.
\newblock Designing spontaneous behavioral switching via chaotic itinerancy.
\newblock {\em Science Advances}, 6(46):eabb3989, 2020.

\bibitem{Bahdanau2014neural}
Dzmitry Bahdanau, Kyunghyun Cho, and Yoshua Bengio.
\newblock Neural machine translation by jointly learning to align and
  translate.
\newblock {\em arXiv}, 1409.0473, 2014.

\bibitem{Vaswani2017attention}
Ashish Vaswani, Noam Shazeer, Niki Parmar, Jakob Uszkoreit, Llion Jones,
  Aidan~N Gomez, \L~ukasz Kaiser, and Illia Polosukhin.
\newblock Attention is all you need.
\newblock In {\em Advances in Neural Information Processing Systems},
  volume~30, 2017.

\bibitem{Han2022survey}
Kai Han, Yunhe Wang, Hanting Chen, Xinghao Chen, Jianyuan Guo, Zhenhua Liu,
  Yehui Tang, An~Xiao, Chunjing Xu, Yixing Xu, Zhaohui Yang, Yiman Zhang, and
  Dacheng Tao.
\newblock A survey on vision transformer.
\newblock {\em IEEE Transactions on Pattern Analysis and Machine Intelligence},
  45(1):87--110, 2023.

\bibitem{Jumper2021highly}
John Jumper, Richard Evans, Alexander Pritzel, Tim Green, Michael Figurnov,
  Olaf Ronneberger, Kathryn Tunyasuvunakool, Russ Bates, Augustin
  {\v{Z}}{\'\i}dek, Anna Potapenko, et~al.
\newblock Highly accurate protein structure prediction with {A}lphafold.
\newblock {\em Nature}, 596(7873):583--589, 2021.

\bibitem{Thiele2018neuromodulation}
Alexander Thiele and Mark~A Bellgrove.
\newblock Neuromodulation of attention.
\newblock {\em Neuron}, 97(4):769--785, 2018.

\bibitem{Edelmann2018dopaminergic}
Elke Edelmann and Volkmar Lessmann.
\newblock Dopaminergic innervation and modulation of hippocampal networks.
\newblock {\em Cell and tissue research}, 373(3):711--727, 2018.

\bibitem{Palancios2019neuro}
Jon Palacios-Filardo and Jack~R Mellor.
\newblock Neuromodulation of hippocampal long-term synaptic plasticity.
\newblock {\em Current Opinion in Neurobiology}, 54:37--43, 2019.

\bibitem{Yildiz2012revisiting}
I.~B. Yildiz, H.~Jaeger, and S.~J. Kiebel.
\newblock Re-visiting the echo state property.
\newblock {\em Neural Networks}, 35:1--9, 2012.

\bibitem{Lukosevicius2012practical}
M.~Luko{\v{s}}evi{\v{c}}ius.
\newblock A practical guide to applying echo state networks.
\newblock {\em Neural Networks: Tricks of the Trade}, pages 659--686, 2012.

\bibitem{Jaeger2012long}
Herbert Jaeger.
\newblock Long short-term memory in echo state networks: Details of a
  simulation study.
\newblock {\em Jacobs University Technical Reports}, 27, 2012.

\bibitem{Inubushi2017reservoir}
Masanobu Inubushi and Kazuyuki Yoshimura.
\newblock Reservoir computing beyond memory-nonlinearity trade-off.
\newblock {\em Scientific reports}, 7(1):10199, 2017.

\bibitem{Jordanou2022investigation}
Jean~Panaioti Jordanou, Eric~Aislan Antonelo, Eduardo Camponogara, and Eduardo
  Gildin.
\newblock Investigation of proper orthogonal decomposition for echo state
  networks.
\newblock {\em arXiv}, 2211.17179, 2022.

\bibitem{Lu2017reservoir}
Zhixin Lu, Jaideep Pathak, Brian Hunt, Michelle Girvan, Roger Brockett, and
  Edward Ott.
\newblock Reservoir observers: Model-free inference of unmeasured variables in
  chaotic systems.
\newblock {\em Chaos: An Interdisciplinary Journal of Nonlinear Science},
  27(4):041102, 2017.

\bibitem{Katori2019reservoir}
Yuichi Katori, Hakaru Tamukoh, and Takashi Morie.
\newblock Reservoir computing based on dynamics of pseudo-billiard system in
  hypercube.
\newblock In {\em 2019 International Joint Conference on Neural Networks
  (IJCNN)}, pages 1--8, 2019.

\bibitem{Inubushi2020transfer}
Masanobu Inubushi and Susumu Goto.
\newblock Transfer learning for nonlinear dynamics and its application to fluid
  turbulence.
\newblock {\em Phys. Rev. E}, 102:043301, Oct 2020.

\bibitem{Poole2016exponential}
Ben Poole, Subhaneil Lahiri, Maithra Raghu, Jascha Sohl-Dickstein, and Surya
  Ganguli.
\newblock Exponential expressivity in deep neural networks through transient
  chaos.
\newblock In {\em Advances in Neural Information Processing Systems},
  volume~29, 2016.

\bibitem{Zhang2020expressivity}
Gege Zhang, Gangwei Li, Weining Shen, and Weidong Zhang.
\newblock The expressivity and training of deep neural networks: Toward the
  edge of chaos?
\newblock {\em Neurocomputing}, 386:8--17, 2020.

\bibitem{Inoue2022transient}
Katsuma Inoue, Soh Ohara, Yasuo Kuniyoshi, and Kohei Nakajima.
\newblock Transient chaos in bidirectional encoder representations from
  transformers.
\newblock {\em Phys. Rev. Res.}, 4:013204, Mar 2022.

\bibitem{Barak2013from}
Omri Barak, David Sussillo, Ranulfo Romo, Misha Tsodyks, and L.F. Abbott.
\newblock From fixed points to chaos: Three models of delayed discrimination.
\newblock {\em Progress in Neurobiology}, 103:214--222, 2013.

\bibitem{Sussilo2013opening}
David Sussillo and Omri Barak.
\newblock {Opening the Black Box: Low-Dimensional Dynamics in High-Dimensional
  Recurrent Neural Networks}.
\newblock {\em Neural Computation}, 25(3):626--649, 03 2013.

\bibitem{Seoane2019evolutionary}
Luís~F. Seoane.
\newblock Evolutionary aspects of reservoir computing.
\newblock {\em Philosophical Transactions of the Royal Society B: Biological
  Sciences}, 374(1774):20180377, 2019.

\bibitem{Suarez2021learning}
Laura~E Su{\'a}rez, Blake~A Richards, Guillaume Lajoie, and Bratislav Misic.
\newblock Learning function from structure in neuromorphic networks.
\newblock {\em Nature Machine Intelligence}, 3(9):771--786, 2021.

\bibitem{Bauer2019real}
F.~C. {Bauer}, D.~R. {Muir}, and G.~{Indiveri}.
\newblock Real-time ultra-low power ecg anomaly detection using an event-driven
  neuromorphic processor.
\newblock {\em IEEE Transactions on Biomedical Circuits and Systems},
  13(6):1575--1582, 2019.

\bibitem{Yamaguchi2019chaotic}
M.~{Yamaguchi}, Y.~{Katori}, D.~{Kamimura}, H.~{Tamukoh}, and T.~{Morie}.
\newblock A chaotic {Boltzmann} machine working as a reservoir and its analog
  {VLSI} implementation.
\newblock In {\em 2019 International Joint Conference on Neural Networks
  (IJCNN)}, pages 1--7, July 2019.

\bibitem{Appeltant2011information}
L.~Appeltant, M.~C. Soriano, G.~Van~der Sande, J.~Danckaert, S.~Massar,
  J.~Dambre, B.~Schrauwen, C.~R. Mirasso, and I.~Fischer.
\newblock Information processing using a single dynamical node as complex
  system.
\newblock {\em Nature communications}, 2:468, 2011.

\bibitem{Brunner2018tutorial}
D.~Brunner, B.~Penkovsky, B.~A. Marquez, M.~Jacquot, I.~Fischer, and L.~Larger.
\newblock Tutorial: Photonic neural networks in delay systems.
\newblock {\em Journal of Applied Physics}, 124(15):152004, 2018.

\bibitem{Penkovsky2018efficient}
Bogdan Penkovsky, Laurent Larger, and Daniel Brunner.
\newblock Efficient design of hardware-enabled reservoir computing in {FPGAs}.
\newblock {\em Journal of Applied Physics}, 124(16):162101, 2018.

\bibitem{Werbos1990backpropagation}
P.~J. {Werbos}.
\newblock Backpropagation through time: what it does and how to do it.
\newblock {\em Proceedings of the IEEE}, 78(10):1550--1560, Oct 1990.

\bibitem{Wright2022deep}
Logan~G Wright, Tatsuhiro Onodera, Martin~M Stein, Tianyu Wang, Darren~T
  Schachter, Zoey Hu, and Peter~L McMahon.
\newblock Deep physical neural networks trained with backpropagation.
\newblock {\em Nature}, 601(7894):549--555, 2022.

\bibitem{Lillicrap2016random}
Timothy~P Lillicrap, Daniel Cownden, Douglas~B Tweed, and Colin~J Akerman.
\newblock Random synaptic feedback weights support error backpropagation for
  deep learning.
\newblock {\em Nature communications}, 7(1):13276, 2016.

\bibitem{Nokland2016direct}
Arild N{\o}kland.
\newblock Direct feedback alignment provides learning in deep neural networks.
\newblock In {\em Advances in Neural Information Processing Systems},
  volume~29, 2016.

\bibitem{Guerguiev2017towards}
Jordan Guerguiev, Timothy~P Lillicrap, and Blake~A Richards.
\newblock Towards deep learning with segregated dendrites.
\newblock {\em eLife}, 6:e22901, dec 2017.

\bibitem{Murray2019local}
James~M Murray.
\newblock Local online learning in recurrent networks with random feedback.
\newblock {\em eLife}, 8:e43299, may 2019.

\bibitem{Frenkel2021learning}
Charlotte Frenkel, Martin Lefebvre, and David Bol.
\newblock Learning without feedback: Fixed random learning signals allow for
  feedforward training of deep neural networks.
\newblock {\em Frontiers in neuroscience}, 15:629892, 2021.

\bibitem{Bellec2020solution}
Guillaume Bellec, Franz Scherr, Anand Subramoney, Elias Hajek, Darjan Salaj,
  Robert Legenstein, and Wolfgang Maass.
\newblock A solution to the learning dilemma for recurrent networks of spiking
  neurons.
\newblock {\em Nature communications}, 11(1):3625, 2020.

\bibitem{Nakajima2022physical}
Mitsumasa Nakajima, Katsuma Inoue, Kenji Tanaka, Yasuo Kuniyoshi, Toshikazu
  Hashimoto, and Kohei Nakajima.
\newblock Physical deep learning with biologically inspired training
  method: gradient-free approach for physical hardware.
\newblock {\em Nature Communications}, 13:7847, 2022.

\bibitem{Kingma2014adam}
Diederik~P. Kingma and Jimmy Ba.
\newblock Adam: A method for stochastic optimization.
\newblock {\em arXiv}, 1412.6980, 2014.

\bibitem{Pazke2019pytorch}
Adam Paszke, Sam Gross, Francisco Massa, Adam Lerer, James Bradbury, Gregory
  Chanan, Trevor Killeen, Zeming Lin, Natalia Gimelshein, Luca Antiga, Alban
  Desmaison, Andreas Kopf, Edward Yang, Zachary DeVito, Martin Raison, Alykhan
  Tejani, Sasank Chilamkurthy, Benoit Steiner, Lu~Fang, Junjie Bai, and Soumith
  Chintala.
\newblock Pytorch: An imperative style, high-performance deep learning library.
\newblock In {\em Advances in Neural Information Processing Systems},
  volume~32, 2019.

\bibitem{Snoek2012practival}
J.~Snoek, H.~Larochelle, and R.~P. Adams.
\newblock Practical bayesian optimization of machine learning algorithms.
\newblock In {\em Advances in Neural Information Processing Systems 25}, pages
  2951--2959. 2012.

\bibitem{Frazier2018tutorial}
P.~I. Frazier.
\newblock A tutorial on bayesian optimization.
\newblock {\em arXiv:1807.02811}, 2018.

\bibitem{Doya1992bifurcations}
K.~Doya.
\newblock Bifurcations in the learning of recurrent neural networks.
\newblock In {\em [Proceedings] 1992 IEEE International Symposium on Circuits
  and Systems}, volume~6, pages 2777--2780 vol.6, 1992.

\bibitem{Pascanu2013on}
Sanjoy Dasgupta and David McAllester, editors.
\newblock {\em On the difficulty of training recurrent neural networks},
  volume~28 of {\em Proceedings of Machine Learning Research}, Atlanta,
  Georgia, USA, 17--19 Jun 2013. PMLR.

\end{thebibliography}

\end{document}